\documentclass{article}

\PassOptionsToPackage{numbers,square}{natbib}
\usepackage[preprint]{neurips_2024}

\usepackage[utf8]{inputenc} 
\usepackage[T1]{fontenc}    
\usepackage{hyperref}       
\usepackage{url}            
\usepackage{booktabs}       
\usepackage{amsfonts}       
\usepackage{nicefrac}       
\usepackage{microtype}      
\usepackage{xcolor}         
\usepackage{graphicx} 
\usepackage[utf8]{inputenc}

\definecolor{mydarkblue}{rgb}{0,0.08,0.45}
\hypersetup{
colorlinks=true,
linkcolor=mydarkblue,
citecolor=mydarkblue,
filecolor=mydarkblue,
urlcolor=mydarkblue,
}

\newcommand{\thead}[1]{\small\bfseries\begin{tabular}{@{}c@{}}#1\end{tabular}}
\usepackage{todonotes}

\usepackage{aliascnt}

\newaliascnt{appendixsubsection}{subsection}

\aliascntresetthe{appendixsubsection}

\usepackage{amsmath}
\usepackage{amssymb}
\usepackage{mathtools}
\usepackage{amsthm}

\title{Verifiable evaluations of machine learning models using zkSNARKs}

\author{%
  Tobin South\thanks{tsouth@mit.edu} \\
  MIT \\
  \And
  Alexander Camuto \\
  EZKL \\
  \And
  Shrey Jain \\
  Microsoft Research \\
  \And
  Shayla Nguyen \\
  University of Adelaide \\
  \And
  Robert Mahari \\
  MIT \\
  \And
  Christian Paquin \\
  Microsoft Research \\
  \And
  Jason Morton \\
  EZKL \\
  \And
  Alex `Sandy' Pentland \\
  MIT \\
}

\begin{document}

\maketitle

\begin{abstract}
In a world of increasing closed-source commercial machine learning models, model evaluations from developers must be taken at face value. These benchmark results---whether over task accuracy, bias evaluations, or safety checks---are traditionally impossible to verify by a model end-user without the costly or impossible process of re-performing the benchmark on black-box model outputs. This work presents a method of verifiable model evaluation using model inference through zkSNARKs. The resulting zero-knowledge computational proofs of model outputs over datasets can be packaged into verifiable evaluation attestations showing that models with fixed private weights achieve stated performance or fairness metrics over public inputs. We present a flexible proving system that enables verifiable attestations to be performed on any standard neural network model with varying compute requirements. For the first time, we demonstrate this across a sample of real-world models and highlight key challenges and design solutions. This presents a new transparency paradigm in the verifiable evaluation of private models. 
\end{abstract}

\section{Introduction}
Model transparency, bias checking, and result reproducibility are at odds with the creation and use of closed-sourced machine learning (ML) models. It is common practice for researchers to release model weights and architectures to provide experimental reproducibility, foster innovation, and iteration, and facilitate model auditing of biases. However, the drive towards commercialization of models by industry (and, in the case of extremely large language models, the concern over the safety of open-source models) has led to the increasing practice of keeping model weights private \citep{karpur2023securing,brown2020language}.

\begin{figure}[th]
    \centering
    \includegraphics[width=\textwidth]{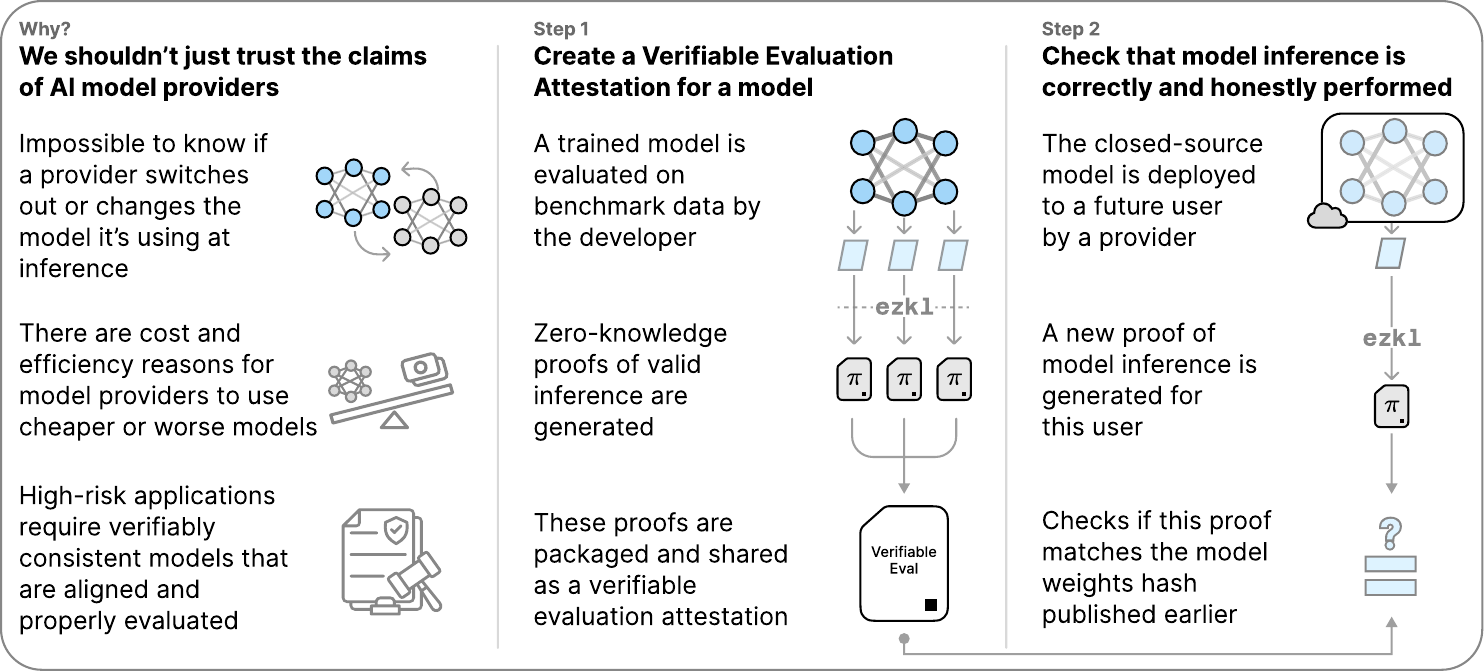}
    \caption{A high level overview of the motivations and system design, which is augmented by the flexible ezkl proving system that can handle any ML model.}
\end{figure}

Keeping model weights private (regardless of whether the model architecture is public) limits the ability of external observers to examine the model and its performance properties. This presents two key concerns. Firstly, it can be hard to verify any claims that are made about a model’s performance, either in the form of scientific evaluation results or commercial marketing. Secondly, performance characteristics that were not intended or tested for are hard to examine. Instances of racial and gendered bias have been found post-hoc in production ML models~\citep{buolamwini2018gender}. These unwanted model performance characteristics, such as bias, come in many forms and are often not initially tested for, leaving it to the public and interested parties to determine the characteristics of a model.

Tools such as algorithmic audits~\citep{buolamwini2018gender, raji2019actionable,kroll2015accountable} can partially allow auditing via API, but come at a significant expense and are not possible when models are not publicly facing. Many high-risk and customer-impacting models exist without public APIs, such as resume screeners~\citep{bogen2018help} or models used internally by law enforcement or governments. Even cases where models are public can benefit from verifiable evaluations given the technically challenging and expensive process of running these models.

While a model creator or provider can always state claims about a specific model’s performance on an evaluation task or set of characteristics (often in the forms of documentation or model cards such as \citet{mitchell2019model}), this is insufficient to verify a model being used has a performance characteristic. To verify the claimed capability, an end user must be able to confirm that the model in use was run on the benchmark in question, that it performs at a desired level, and that the same model is being used at a later date. This problem is not purely academic, as some scholars have raised concerns that OpenAI's models have degraded over time~\citep{chen2023chatgpt}.

\paragraph{The problem.} Users of AI systems have a vested interest in knowing about the performance capabilities of these systems. At the same time, high-performance AI systems may be more expensive to provide, or systems that perform well on certain benchmarks may underperform elsewhere. This creates incentives for model providers to provide users with a different model than advertised. Users without technical sophistication or computational resources struggle to confirm that the models they are using meet advertised benchmarks. While the EU AI Act and the US Executive Order on AI include transparency obligations for model providers to share technical details about the model \citep{us2023executive, eu2023aiact}, enforcing compliance with this requirement is extremely challenging in practice (especially as the market for AI becomes increasingly fragmented). We sketch a technical safeguard against this practical problem by allowing AI system users to verify the properties of the model they are using without requiring them to run their own tests or trust model providers.

\paragraph{Contributions.} To address these challenges, this work presents the first end-to-end demonstration of using zero-knowledge machine learning to generate evaluation proofs on arbitrary ML models. 
In doing so, this work contributes: \vspace{-1em}
\begin{itemize} \setlength\itemsep{-0.2em}
\item A framework for general-purpose verifiable attestation of neural networks and other models across a variety of datasets and tasks using fully succinct non-interactive proofs;
\item Built upon a new proving approach that maps from onnx model formats to proof circuits, flexible to any ML model.
\item A challenge-based model for checking model inference matches performance-attested model weights.
\item Analysis of computational costs and complexity of model evaluation, with an eye to how speed and scalability can be addressed via a `predict, then prove' strategy. 
\item A demonstration of evaluation across traditional non-neural ML models, multi-layer perceptions, convolutional neural networks, long short-term memory networks, and small transformers. 
\item A working implementation for public use with arbitrary models using the ezkl toolkit.
\end{itemize}

\section{Background and Related Works}
This work seeks to combine the extensive literature in accountable algorithms \citep{kroll2015accountable,kim2017auditing,desai2017trust}, ML reproducibility \citep{Beam2020ChallengesTT,semmelrock2023reproducibility,Albertoni2023ReproducibilityOM,Belz2023MissingIU,10.5555/3546258.3546422,10.1001/jama.2019.20866,10.1145/3576915.3623130,gundersen2022machine} and AI fairness \citep{Muthukumar2018UnderstandingUG,Mehrabi2019ASO,CorbettDavies2018TheMA,May2019OnMS,Raji2019ActionableAI,Krkkinen2021FairFaceFA,Liu2023FairCompassOF,Parraga2023FairnessID} with the growing field of zero-knowledge machine learning \citep{Feng2021ZENAO,Liu2021zkCNNZK,Lee2020vCNNVC,kang2022scaling}, in a pragmatic way. Previous work has typically focused on model-specific proofs, unlike our flexible proving system. Recently, speed improvements have enabled proofs for small language models \citep{sun2024zkllm, chen2024zkml} of similar size to our work, prompting questions of verifiable evaluation of models. While previous works briefly discuss this \cite{kang2022scaling, chen2024zkml}, we flesh out the practical constraints and design choices involved in the verifiable evaluation of models and provide possible solutions to scalability beyond faster chips and future research. A broader and more comprehensive background and related work is available in \autoref{sec:background}.

\section{Problem Definition and Threat Model} \label{sec:problemdef}
A model provider has a private model, $f(\cdot, W)$, with weights $W$. The provider wants to make a claim, $C$, about the model's performance over a labeled dataset $\{(x_i, y_i) | i \in {0, \ldots, N}\}$. Traditionally, the provider would publish the model outputs $\{(x_i, f(x_i)) | i \in {0, \ldots, N}\}$ or an aggregation of those outputs (e.g., $ \frac{1}{N} \sum_0^N \left|y_i-f(x_i)\right|$) (e.g., a benchmark). If a model provider is a bad actor, they can simply produce fake outputs or aggregation statistics (i.e., lie about the model performance). 

Here, we want the ability to verify these published facts without needing the full weights. To do so, we publish the results and an attached proof of inference $\pi$ containing the model weights hash $H(W)$. The output then becomes $\{(x_i, f(x_i), \pi, H(W)) | i \in {0, \ldots, N}\}$ or an aggregated claim such as $ (\Pi, H(W), \frac{1}{N} \sum_0^N \left|y_i-f(x_i)\right|)$ where $\Pi$ is a meta-proof that verifies and attests to all included inferences being valid and the aggregation step being correctly computed. It must be possible to verify that the inferences are valid and that the model weight hash matches those inferences.

This is sufficient for scientific result confirmation but insufficient for models being deployed in the real world. To confirm the future use of a model is done by the same model with matching performance characteristics, a proof of inference must be done. This can be achieved by sharing with an end user the result of the inference on their input $x^*$ as well as a proof of inference and the model weights hash $(x^*, f(x^*), \pi, H(W))$. The data that is fed into this proof is called the `witness.' This allows the end user to confirm the model they access matches the model of the performance claim by verifying that the proof is valid and the corresponding model weight hash matches the original claim. A full threat model is available in \autoref{sec:threatmodel}.

\subsection{Example Uses}
To further identify the need for this system, we outline two simple real-world threats where verifiable evaluations are needed. First, consider the case of a new model architecture with private weights being published in an academic context, where the authors claim a high performance on a benchmark. As the model weights are private, model users and auditors cannot verify that the model performs at the described levels. The public needs to verify that a model \emph{exists} such that its execution over a dataset produces the benchmark result. 
The verifiable evaluation performance attestation demonstrates that the authors have a set of model weights that can achieve such performance on an evaluation. Even though weights are kept private, the general architecture of the model (e.g., whether it is a CNN or transformer) will be implicitly leaked in the current proof system (although future work could obfuscate some elements of this such as the number of layers in a CNN at the cost of proof size).

Second, consider the case of a model being used in production, via a publicly facing API or for internal use only. As above, a model consumer wants to know that there exists a model that performs well and has the described characteristics (such as not exhibiting racial bias on test sets). However, the model consumer also wants to know that this well-performant model is the one being used during inference. This motivates the model weight hash verification during inference time. 

\subsubsection{Scalability in production}
Performing zero-knowledge proof of inference is a computationally expensive process compared to standard ML inference. Previous work treats the steps of inference and proving as intertwined \citep{kang2022scaling, chen2024zkml}; when they don't need to be. The zkSNARK computation operates on a witness, $(x^*, f(x^*), H(W))$, which must be generated at inference time with very low compute cost, but the proof itself can come later. This, in essence, leverages the fact that zkSNARKs prove that a witness and a circuit combine to produce valid outputs.

In high-risk contexts, an inference proof can be generated and checked for every run \emph{before} using model outputs (as suggested in previous work). However, it is possible to use a `predict, then prove' strategy, in which the model inference is provided immediately, but the proof (possibly computed in parallel) is provided at a delay after the proof process completes. This delay could be seconds, minutes, or more and still provide useful finality that doesn't slow down the delivery of the result. This is especially relevant in legal and regulatory regimes where model performance needs to remain auditable over time, including contexts such as using FDA-approved models in medical settings or model vendors used in high-risk settings (akin to `Trust, but Verify' \citep{desai2017trust}). 

Alternatively, model consumers can choose, for a cost, to challenge the model provider to present a proof of inference and the model weight hash for an input of concern. This guarantees only this inference, not all previous inferences, but can be leveraged strategically with a much lower total burden. If challenges are done randomly, a probabilistic guarantee of the model is created; or challenges can be done when model outputs are suspicious. Conditional on randomness in inference and reasonable accuracy bounds due to quantization, an audit can be performed post hoc on an inference output pair so long as a verifiable evaluation attestation and witness exist. 
A longer discussion of challenge-based model auditing can be found in  \autoref{sec:challenge}.


\section{System Design}

\begin{figure*}[thpb]
    \centering
    \includegraphics[width=0.9\textwidth]{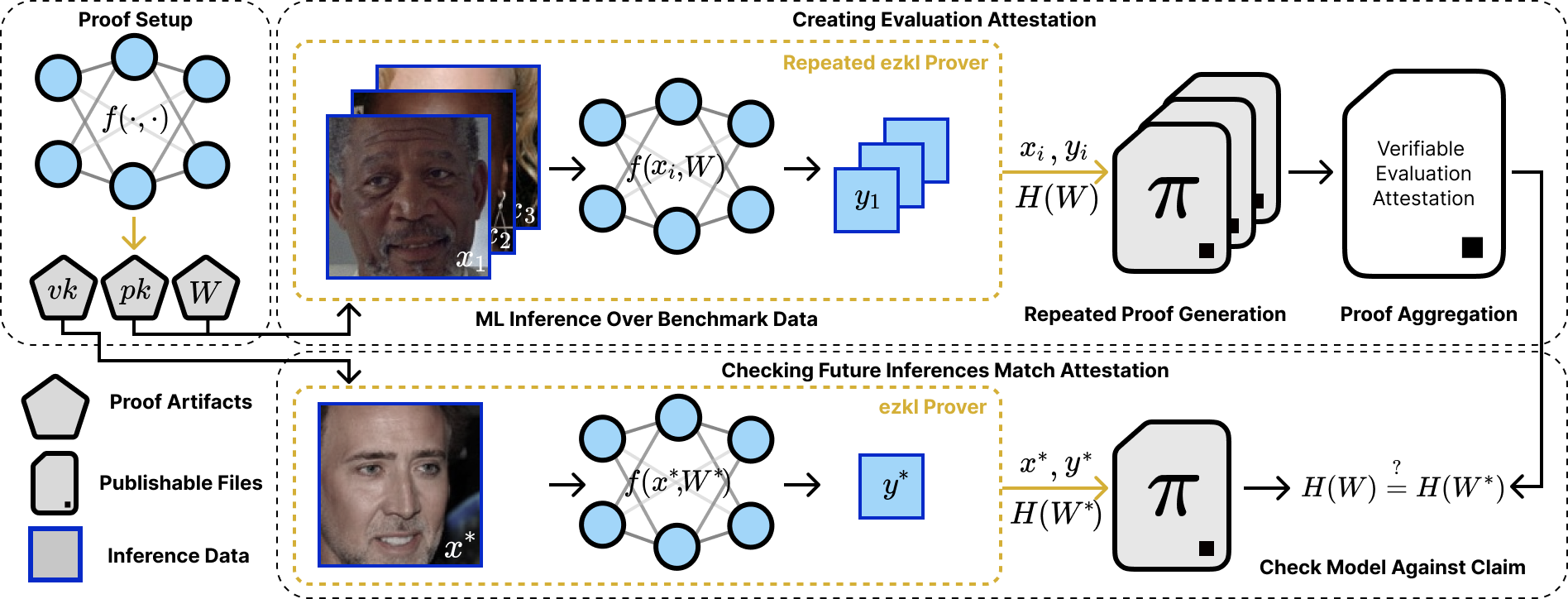}
    \caption{System diagram of verifiable ML evaluation using the zkSNARK ezkl toolkit. A model can be compiled into a proving key ($pk$) and verification key ($vk$) which can be used to generate repeated inference proofs over a dataset ($\pi$), which can then be aggregated into a verifiable evaluation attestation. Using the same proving and verification keys, any future inference of a model can be checked to confirm a model with the same model weight hash, $H(W)$, was used to generate the output. Inference data can be arbitrary.
    }
    \label{fig:fullsystem}
\end{figure*}

To generate and audit verifiable evaluation attestations, the system follows a broad four-step process shown in \autoref{fig:fullsystem}. 
First, a model of interest that has been trained and prepared for deployment is set up for inference. The model is converted into a standard model format (ONNX) and a circuit corresponding to the internal operations of the inference is created. This circuit generation process can be calibrated according to accuracy vs. resource tradeoffs and generates a large proving key, $pk$, and a verification key, $vk$, for the proof setup. This proof setup takes the form $\text{Setup}(1^{\lambda}, W, f) \rightarrow (pk,vk)$ using a standard security parameter $\lambda$, the model weights, and architecture to produce a commitment to the execution of a circuit. This circuit corresponds to the sequence of operations used in the inference of the model. This work is the first to present a flexible setup that works for any ML model in ONNX format to be converted to a proof circuit, made possible by creating a large number of interoperable proof arguments to support the diversity of model architectures; technical details of how this was achieved are in \autoref{sec:setup}.


Second, an evaluation dataset is selected to confirm performance or check for bias. Inference over the dataset is performed using standard practices to produce a set of input files containing input-output pairs $(x,y)$. Dataset choice is critical and can be dynamic (e.g., red teaming) as explored in \autoref{sec:datasetchoice}.

Third, for each data pair, a witness file is generated (including some quantization) for the $(x,y)$ pair, and a computational inference proof is created as a zkSNARK. This proof takes $W$ as a secret input value to the circuit and $x$ as a public input value and uses the previous commitment $pk$ to generate the proof $\pi$. This $\text{Prove}(pk, W, x, y) \rightarrow \pi \supset \left\{H(W), y\right\} $ is a computationally expensive step, with more details in \autoref{sec:datasetinference}.

These proof files can be aggregated into a verifiable evaluation attestation, including artifacts showing model performance and consistent model weight hashes. This can either be done by naively bundling the inference proofs or through an additional aggregation circuit that checks if all the proofs are valid and combines all the outputs into one. This additional setup comes with important additional privacy considerations for the model results and is elaborated in \autoref{sec:aggregation}. This claim can be publicly published to the world without the risk of leaking model weights.

At a later time, a model user or auditor who sees the result of a model inference $(x^*, y^*)$ can challenge the model provider to confirm which model was used to generate this pair. At an increased cost over standard inference (levied on either the auditor or provider), the provider holding the private model weights can be asked to prove using a new zkSNARK that such an input-output pair can be generated within accuracy bounds from a model with a matching circuit and model weight hash. By performing $\text{Prove}(pk, W, x^*, y^*) \rightarrow \pi^* $ using the same $W$ and $pk$, the provider can prove that they can generate $y^*$ using the model with $H(W)$, and hence the model with the same benchmark properties from the verifiable evaluation attestation. If a proof cannot be generated with matching hashes, the challenge has been failed. More discussion of this challenge-based approach can be found in \autoref{sec:challenge}.

\subsection{Aggregation of proofs}\label{sec:aggregation}

Once a collection of inference proofs has been generated, there is a question of how to package them for publication and sharing. The simplest naive solution is to zip up the proofs, which include the witness outputs, and a verification key, allowing any future examiner to calculate performance measures. This is computationally efficient and allows for flexible measures of performance to be made. However, this naive bundle exposes the test data outputs to unwanted use (such as training a copycat model based on these outputs \citep{hinton2015distilling}). Even though this is a reduced risk on test benchmarks, some model providers would prefer not to release inference.

Having the full test data inference results available also affords the possibility of a post-hoc audit on other properties contained in that data (e.g., a test set might only be intended for showing performance, but a careful audit of response performance across classes could reveal biases in the performance across classes). One such example might be sharing overall facial recognition accuracy, of which a finer grain analysis would show that the performance is much larger for white and male faces~\citep{buolamwini2018gender}. Since proof files and verification keys are very small, these naive bundles of inference proofs remain small and fast to verify even for large dataset sizes, with the vast majority of the work in proof generation.

A secondary approach would be to perform a vanilla aggregation circuit which checks if all the proofs are valid and combines all the outputs into one. This has similar downsides to compressing the collection but now comes with the additional cost of the aggregation circuit, which can become quite large. This zero-knowledge aggregation step makes sense when posting model inferences to a blockchain, but less so when sharing a performance attestation. We provide an implementation of this through the aggregation toolkit in ezkl.

A third approach uses such an aggregation circuit but explicitly creates a custom Halo2 circuit to aggregate just the result we want (e.g., the accuracy score, confusion matrix). This allows for a separate privacy regime to be applied, where data outputs are kept private, and only a verifiable accuracy metric is shared. This additional step adds complexity both in the form of additional computational requirements and the need to audit accuracy calculation code, but adds a significant layer of privacy. This additional zero-knowledge proof (ZKP) approach would need to be updated for different model types and outputs (e.g., models with varying input output formats, models with different accuracy measures, or benchmarks requiring complex generation of evaluation like the HumanEval example in \autoref{sec:examplemodels}).

\section{Experimental Results}

To demonstrate the flexibility and utility of the verifiable model attestations, we perform two groups of experiments. First, we show an evaluation of example models from a range of modalities, ranging from bias checking of facial detection via standard convolutional neural networks to limited inference of a language model. Second, we perform inference on a series of network architectures at varying model sizes to estimate total memory and compute time requirements, so as to estimate total costs of benchmarking. Code to rerun experiments and use the system can be found on Github\footnote{
\url{https://github.com/<REDACTED>} (or look for the zip file in SI)}.

\subsection{Example Verifiable Model Attestations}\label{sec:examplemodels}

While general in design, the specifics of generating a verifiable evaluation attestation vary across model types. Any model that can be expressed as a computational graph in ONNX can be evaluated, and a sample of example models is shown in \autoref{tab:examplemodels}.

The simplest such example of an evaluation would be a multi-layer perceptron (MLP) benchmarking on a small dataset such as MNIST \citep{deng2012mnist}. 
An inference proof is generated on each flattened image from the held-out test split to show its mapping from the input to a number. Bundling these inference proofs with the model weight hash and the ground truth values allows an inspector to evaluate the model accuracy by simply summing across the witness outputs and the ground truths in the attestation. Given the openness of the dataset, there is no need to privatize the ground truths. However, the naive bundling of the inference proofs for an attestation allows inspectors to see exact model outputs. Instead, a simple zero-knowledge proof (ZKP) could be made to verify each inference with the verification key, confirm $H(W)$ remains constant, and perform the simple $\tilde{y}_i \mathrel{\overset{\scriptscriptstyle ?}{\scriptstyle =}} y_i$ comparison across all the test data to produce a verifiable evaluation attestation without releasing individual data. In this case, the information in the naive bundle, including the ground truths are provided as witnesses to the proof with model outputs as private inputs and ground truths as public inputs. The full attestation with all test data proofs and a verification key is less than 100MB.


Other small models can behave exactly the same. A benchmark over a CNN applied to MNIST can be done exactly the same (with different performance characteristics as explored below). Models such as linear regression, SVM, or random forest can all be benchmarked in this approach. While these are not natively convertible to ONNX, there are a variety of tools to convert standard sklearn models into tensor-based models for use on ONNX such as hummingbird \citep{nakandala2019compiling} and sk2torch. We use a small dataset \citep{bennett1992robust} to demonstrate their use due to their small size.

Models can be more than classifiers. Consider a variational autoencoder (VAE) \citep{kingma2014auto} trained on the CelebA faces dataset \citep{liu2015deep}. While full inference of VAE can be proved, the internal reparameterization can create minor challenges for proofs due to the random generation. Instead we prove a partial execution of the VAE showing decoding from the latent space. Here the witness inputs are points in the latent space to be passed into the decoder and a naive bundle will contain these points with proven image outputs from the latent space. This provides an interesting approach to allowing for verifiable statements about the latent space. So long as the space has been sufficiently sampled (such as via a grid across the space centered around the mean) an inspector could see if any undesirable outputs are generated or inspect to see the degree to which certain properties of the latent space (such as race in facial images) are clustered. 

This generative approach can be applied to small language models as well. Autoregressive models such as an LSTM~\citep{sak2014long} or a decoder-only transformer~\citep{radford2019language} require a proof for each inference (e.g., each token they generate). This makes the computational cost of generating a long sequence of tokens quite high. It’s worth noting, however, that these proofs can be parallelized. Since the witness can be generated preemptively before proving, the full sequence of tokens can be generated from normal inference to create a series of witnesses that autoregressively show the next tokens being generated, which can then each be proven in parallel. We demonstrate small language models (nanoGPT\footnote{\url{https://github.com/karpathy/nanoGPT}}, designed to replicate GPT-2) running on a sample of HumanEval~\citep{chen2021evaluating}. These language models can be pretrained as the verifiable benchmarking attestation only requires inference.

We assume a public tokenizer and pass pre-tokenized sequences of tokens from the prompt into the model to generate the next token. For simple benchmarking datasets with a single token answer from multiple choice reasoning (such as MMLU \citep{hendryckstest2021}) this can be straightforward to benchmark and package. However, the space of LLMs has opened more rich and complex benchmarks such as HumanEval~\citep{chen2021evaluating} which requires generating longer strings as code and then evaluating the execution that code. Creating a naive bundle as a verifiable evaluation attestation is simple, requiring only that all token generation steps be put together such that an inspector can confirm that the code was generated by the model with the hash $H(W)$. Generating a zero-knowledge proof to privatize the generated code while attesting to its performance is much harder. This would require not only verifying each inference using the verification key, but also doing so in the correct order ensuring coherence of the previous tokens in the witness, and verifying the correct execution of the code and its output.

These approaches can be extended to a variety of models, such as other autogressive models or new language model architecture, diffusion models which behave much like VAE decoders, or sequence to sequence models such as Wav2Vec~\citep{baevski2020wav2vec} or Whisper~\citep{fedus2022switch} which apply the public tokenizer assumption to audio processing and generate a sequence output as witness. Larger models, such as a mixture of experts, are also possible, with model size being the only constraint. As we see below, proof time and resource requirements grow dramatically with large models. 

\begin{table}
\centering
\caption{Total resource and time requirements for generating an inference proof for example models.
}
\label{tab:examplemodels}
\begin{tabular}{lccccclcc}
\thead{Model}              & \thead{Model\\Params\\($\times 10^3$)}& \thead{Model\\Flops\\($\times 10^3$)
 }& \thead{Constraints\\($\times 10^3$)} & \thead{Prove\\Time (s)} & \thead{Verify\\Time (s)} &\thead{Proof\\Size} & \thead{PK\\Size} & \thead{VK\\Size} \\
\hline
Regression         &              0.03&             --& 0.062& 0.1& 0.01&13K& 715K& 1.7K\\
SVM                &              0.03&             --& 0.626& 0.3& 0.02&23K& 16M& 2.5K\\
Random Forest      &              0.08&             --& 3.627& 2.9& 0.02&26K& 276M& 2.7K\\
MLP                & 3.6& 3.5& 1.920& 0.3& 0.02&21K& 14M& 2.3K\\
Small CNN          & 19.8& 68.6& 35.84& 3.1& 0.03&15K& 390M& 1.8K\\
VAE (decoder)      & 1065& 12582& 2016.6& 142& 0.42&1.9M& 16G& 2.5K\\
LSTM               & 29& 950& 495.7& 35& 0.10&41K& 4.1G& 2.5K\\
nanoGPT & 250& 51396&                   9398.9&              2781&                   2.69&0.7M&         219G&         4.2K\\
\end{tabular}
\end{table}

\subsection{Increasing Costs from Increasing Size}\label{sec:increasingsize}

\begin{figure}[htbp]
    \centering
    \includegraphics[width=\linewidth]{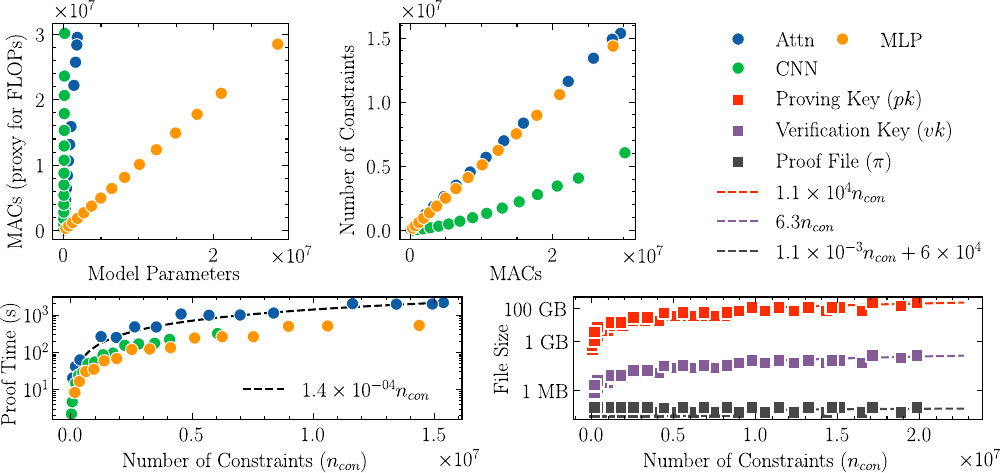}
    \caption{Time and RAM requirements for model proofs with increasing model sizes across multi-layered perceptions (MLP), convolutional neural networks (CNN), and attention-based transformers (Attn). Model requirements scale linearly with the number of constraints, driven by the number of operations used in a model inference. }\label{fig:scaling}
\end{figure}

Increasing numbers of model parameters can lead to dramatically increased performance~\citep{Kaplan2020ScalingLF}. These larger models also require greater resource requirements for inference. The same is true when performing inference of models inside zkSNARKs. 

Increasing the number of operations a model performs---often measured in multiply-accumulate operations (MACs) or floating-point operations (FLOPs)---increases linearly the number of constraints that need to be proved. While the total number of parameters in a model helps determine the number of operations, this is not a direct relationship. Different model architectures use parameters differently, and hence some models (such as CNNs) have a far greater number of operations per parameter than others (such as linear layers in deep neural networks). This relationship can be seen in \autoref{fig:scaling}. 
The CNNs have a lower scaling ratio as optimizations have been made to remove irrelevant operations (such as kernels operating on zero-values) which are included in pytorch operation counts but not in the number of constraints.

To generate an inference proof, the prover must make a KZG commitment~\citep{kzg10} to all the operations needed for inference requiring an SRS file \citep{nikolaenko2022powers} (see \label{sec:trustedsetup}). 
These files and the resulting proving keys they enable can become very large on the order of multiple gigabytes for reasonably sized models to hundreds of gigabytes for extremely large models. In general, this large proving key size and the memory requirements it puts on a proving system are the limitations on model size. 

While these keys can be quite large, the burden is on the prover. The resulting proof sizes and verification keys are extremely small. These too scale linearly with model operations but at a far slower rate, allowing for a large number of proof files and their verification key to be shared without a large file size burden. In turn, the verification time is also extremely quick. 

\section{Discussion \& Limitations}

While both flexible and powerful, this system is still extremely slow. As a result, verifiable evaluations on datasets are expensive. This motivates key classes of questions for discussion. The first centers around how to estimate current evaluation costs and what future research may provide to reduce them, discussed below.
Second, given limited resources important choices must be made over what benchmarks to run, a key question given the challenges in evaluating models. 
Third, a technical discussion of how speed improvements could be achieved is left for \autoref{sec:speedup}.
Finally, we turn attention to private evaluations. These limitations restrict the current use of the system from frontier models, but future work may enable this. Security limitations related to the use of zkSNARKs are outlined in the threat model in \autoref{sec:threatmodel}, as well as the audit limitations of a challenge-based model.

\subsection{Size and Cost Analysis}
For a given model and dataset, the total time it would take to perform inference proofs over the dataset and aggregate it can be roughly approximated. Each specific model architecture has a different relationship between its parameters and the number of operations it performs during inference. However, once a model is profiled for the number of operations (and hence the number of constraints), which can be done extremely efficiently, estimating the total cost to benchmark the model on a given dataset is easy. 
The proof time complexity for a given model is determined by the number of constraints needed to prove. For a given number of constraints, $n_{con}$, the proof rounds up to the nearest SRS file row count, which provides a proof with time complexity of $O\left( 2^{\lceil{\log_2(n_{con})}\rceil}\right)$ which for practical purposes is approximately linear $O(n_{con})$. While the relationship between parameters, operations (MACs), and constraints is usually linear, stating the complexity as $O(n_{parameters})$ masks the varied scaling rate between parameters and constraints across different model types. 

During experimentation, we found proving key size ($pk$) to be the largest limiting factor, even though it grows $O(n_{con})$, as large models require larger memory space to hold the proving key, increasing the size and cost of the compute cluster used.  
To estimate the total compute cost needed after determining the number of constraints for a model, the estimated proving time can be multiplied by the size of datasets to get total runtime, which can be cost calculated for whatever compute cluster has sufficient RAM to hold the proving key. 
Beyond the construction of the verifiable evaluation attestation, these cost estimates are useful for designing costs and incentive structures for model challenges. If challenging a model is costly, either the challenger or provider (or perhaps the loser of the challenger) will need to bear the cost. Having a publicly known cost for model inference makes this process fairer.

\subsection{Dataset Selection Choice}\label{sec:datasetchoice}

A key motivation for this system is the desire to bring transparency, accountability, and robustness to model benchmarking and its importance for fairness and bias checking. This fundamentally relies on our ability to have robust benchmarks to assess model qualities. 
Evaluating ML models is both critically important and often challenging \citep{Liao2021AreWL,HariniFramework}, especially in the case of NLP \citep{jones1995evaluating,kiela2021dynabench,bowman2021will,raji2021ai} and the general purpose usage of foundation models \citep{liang2022holistic,chang2024survey}. 

Even in cases when benchmarking is done well, there are many edge cases that are often missing (e.g., intersectional bias \citep{lalor2022benchmarkingIntersectional,NEURIPS2021_1531beb7,foulds2020intersectional,tan2019assessing} or out-of-domain usage). Measuring and mitigating these issues is, and will continue to be, a challenge in the ML field \citep{blodgett2020language, shah2019predictive, solaiman2023evaluating, bender2021dangers}. While recent work has proposed new more comprehensive benchmarks \citep{liang2022holistic}, further work will be needed for more robust and complete benchmarks; and addressing the challenges of bias in models is more complicated, requiring changes to values and norms around how AI is built and deployed \citep{Schwartz2022TowardsAS, HariniFramework, kroll2015accountable}. 
This is underscored by the push towards prioritizing AI safety \citep{us2023executive}, which includes comprehensive model evaluations. Tasks such as red-teaming model performance or other alignment checks could also be included in a verifiable evaluation attestation; however, for now, a verifiable evaluation attestation is only as useful as the evaluation itself.

\subsection{Private Benchmarks}

In the system design used here, the \emph{choice} of benchmark has been made public. While the exact outputs of the model over a benchmark can be hidden through aggregation (see \autoref{sec:aggregation} for more details), the choice of the dataset is fundamental to knowing what the model performance attests to. However, one could imagine a scenario where a model provider wants to prove their performance over a private benchmark dataset. This is immediately possible by making model inputs and outputs private after aggregation of inference proofs, but it requires additional work to justify, as these results are information-free in a vacuum.

Two models could be compared on private benchmarks, where a hash of a benchmark dataset is shared without ever sharing the model. This could allow a model provider to show improvements in performance over time with new models without ever sharing the benchmark that they use. Similarly, we could have benchmarks for general foundation model performance that are not released to the internet to avoid training on the test, which could position such a system run by a company or regulator as a gold standard of performance measures.
Alternatively, you may want to have a benchmark on a private dataset but with additional abilities to prove facts about the dataset. For example, a verifiable evaluation attestation could be made, including a dataset hash, and various facts about the dataset and its properties could be shared via separate zkSNARKs that include the dataset hash. 

\section{Conclusion}

We present a novel method for verifiable performance and bias benchmarking of ML models using zkSNARKs. This approach addresses the critical issue of verifying model performance claims in environments where model weights are kept private, which is increasingly common in commercial ML applications. This helps ensure transparency and accountability in model evaluations, particularly in high-stakes scenarios where model reliability and fairness are paramount. The system packages together repeated model inference proofs to demonstrate the accuracy of models either through simple bundling of small proofs and verification files or through meta-proofs of performance over model inference proofs. The system's flexibility was demonstrated across a range of ML models ranging from small perceptrons and regression models to medium-sized transformers. Leveraging a `predict, then prove' approach to serving results and proofs combined with a user challenge model of auditing responses reduces the computational costs in production, and shifts compute demands to model trainers. This is the first practical implementation of a verifiable evaluation for arbitrary ML systems, maintaining model weight confidentiality while ensuring model integrity. In doing so, this lays the groundwork for a more transparent and verifiable future for ML model evaluations.

\bibliography{refs}
\bibliographystyle{unsrtnat}


\appendix


\newpage

\section{Detailed Threat Model} \label{sec:threatmodel}
Based on the setup outlined in \autoref{sec:problemdef}, we can more clearly define a trust model for these verifiable evaluation attestations. We consider only two participants here: the model provider and the model user, where we assume the possibility that the model provider is adversarial. While other actors in the AI supply chain exist (e.g., model developers training weights or external auditors who verify on behalf of a user), we focus on those hosting the model and viewing its outputs. 

The model provider, as the adversary, has one of two goals. The first (threat model \#1) is to make a public claim that a model exists with a performance characteristic that is not true (e.g., I have a model that gets 100\% on ImageNet). The second and more dangerous (threat model \#2) is the provider changes what model is being served to an end user from inside their black box model hosting. Here, you can assume the provider is hosting a model in the cloud and serving a user over standard via API (as is standard).

For threat model \#1, only a closed-source model makes sense, as any widely accessible open-source model could be replicated and checked. For threat model \#2, the model provider can be serving either an open-source or closed-source model, but we focus primarily on the closed-source model here. It is still possible to change which model is being served when it is open-source, and it is an important problem to verify the model in use. While this system supports this, the privacy guarantees for model weights create unnecessary overhead when a simple computation trace would suffice. The ezkl toolkit does have an option for removing model weight privacy and hashing, which can speed up proofs and still enable any ML model to be used (unlike existing work). Instead, we focus on closed-weight models, where the privacy of the weights must be maintained while preventing the adversary from switching models at inference.  

For threat model \#1, there are many ways an adversary can achieve its goal without attached proof, the most obvious of which is to state a benchmark performance number alone. For threat model \#2, the complexity of the attack is similarly low; a model host simply needs to point the API endpoint at a similar enough model in the hosting stack that could plausibly behave close to the original. There are many reasons a usually trustworthy model provider may do this, from reducing inference costs with smaller models to shifting to models with more favorable performance characteristics.
To help imagine this, consider a Volkswagen scenario for an AI model, where a vendor fine-tunes a model with safety and bias in mind at the cost of performance, reports these safety results, and then serves the model that was not fine-tuned to get better performance. 

Threat model \#2 requires both a claim of model performance and a new proof during inference to check the model weight hash matches. Under our threat model, we only create a guarantee that the model provided to you to produce an output has the same set of model weights as the model of the verifiable evaluation attestation. 

The goal of this work is to remove the need for the public or an end user to trust the model provider. The zkSNARKs enable verification that computational work with a model with weights $H(W)$ occurred, that it produced a given benchmark, and that it was used for a specific inference that is challenged. This requires no specific hardware assumptions and draws from the assumptions baked into zkSNARKs (see below).

\subsection{Proving later}
We need proof to guarantee that a model will be served. However, proving is slower than regular inference. As a result, real-time verification is unlikely. Instead, we propose two models (outlined in greater detail in \autoref{sec:challenge}). Either some model inferences are challenged (either when performance degrades or at a regular random interval), or all inferences are challenged. Both of these acts leverage the fact that a witness file (the $(x,y,H(W))$ input to the proof) can be generated at almost no cost during inference, and a proof can be done after the fact on this witness. This ever so slightly weakens the guarantee for threat model \#2. Specifically, this creates the guarantee that the provider has a model with $H(W)$ that can produce $y$ from $x$, not that it necessarily served it earlier. However, if proofs are generated for enough diverse data in production, then we asymptotically have the guarantee that the provider is serving a model with exactly the same properties as the model with weights $W$, which for practical purposes is a reasonable guarantee that they are serving the claimed model. 

\subsection{Security Properties}
The security properties of this system derive from the in-built properties provided by zkSNARKs and apply across both threat models. Specifically, the zkSNARKs provide guarantees on: correctness, that the attestation and inference check accurately represent a set of computations that relate to the inputs and outputs of the circuit $(x,y,H(W))$; soundness, that the outputs of the zkSNARK cannot be generated without knowledge of the weights to generate the hash and the output even with a dishonest party; confidentiality of the model weights, which are kept privacy even when at attestation is shared publicly; integrity, ensuring that the model weights or inputs cannot be modified by a dishonest party during a proof; non-repudiation, ensuring that once proofs are published, the model provider cannot deny the authenticity of the published results; succinctness, that proofs and attestations are small in size; and non-interactivity, that a verifier does not need to interact with the prover during verification.

\subsection{Security Assumptions}
Similarly, the zkSNARK approach (built atop the halo2 proving system) comes with the same security assumptions of zkSNARKs themselves with additional carveouts. For the cryptographic assumptions of the prover, we point to \citet{groth2018updatable} and \citet{halo2book}. The key one to highlight is the need for a trusted setup to provide input randomness into the zkSNARK, via a structured reference string (SRS) (or common reference string). This challenge has been faced by a variety of zkSNARK projects, and consequently, a number of solutions exist.

\label{sec:trustedsetup}
\paragraph{Trusted setup} To use a zk-SNARK, a Structured Reference String (SRS), must be created to provide a public set of parameters. The creation of these parameters also generates `toxic waste,' which can reveal the random inputs to the model and remove the security and privacy provided by the SNARK. The ezkl proving system, drawing from the original halo2, uses the Perpetual Powers of Tau ceremony~\citep{nikolaenko2024powers}, which has many ongoing contributors generating an SRS, such that as long as a single contributor can be trusted to throw away their toxic waste, the SRS is secure.

\subsection{Security Limitations}
While we leverage the security limitation of zkSNARKs as a foundation, we introduce two key additional limitations here. The first is that the guarantees provided, that a model exists, is being used, and has properties, are tied to a specific set of evaluations. As noted in \autoref{sec:datasetchoice}, evaluations are often fraught. A model provider could train a model specific to game a set of benchmarks and evaluations and then faithfully serve you this model. The model would verify and pass all checks but be fundamentally bad as the model developer `gamed' the system by overfitting on the benchmarks or evaluations. The second is that the slow proof times limit the utility in synchronous contexts for large models. While we propose the `predict, then prove' approach as a workaround (explained in the main text and elaborated below), it minimizes the real-time security of the system.


\subsection{Challenge-based Production Model Audits}
\label{sec:challenge}

Any inference of a model that produces a visible input-output pair $(x^*, y^*)$ can be challenged. This includes regularly scheduled audits where an inference step is accompanied by an inference proof challenge, as well as post-hoc audits, where an inference that seems suspicious can be checked against an existing verifiable evaluation attestation. The latter approach limits the ability of the model provider to anticipate a challenge, creating incentives to always use the claimed model. A model user could choose to challenge a provider randomly, creating a predictable likelihood of bad actors getting caught, or selectively in cases where a user is dissatisfied with performance.

The key limitation is that each verifiable inference is vastly more expensive than standard inference. In cases where audits are legally mandated (such as in high risk contexts), a model provider could bear this cost. However, in production deployment models, the provider has an incentive to perform as few verifiable inferences as possible. As such, the model user could pay an additional fee for auditing to account for the cost. If an audit passes, the user is satisfied. However, to make auditing worthwhile in cases of concern, a reward system could be put in place, where a user is rewarded if an audit fails. This could be done using escrow or smart contracts, or via simple terms of service in API usage deals. In the case of a random audit with probability $p$ at cost $c$ per verifiable inference, a reward of $\frac{c}{p}$ would cover the costs in expectation for the user with an untrusted model provider.

Further, even a new verifiable evaluation attestation can be created post-hoc. If concerns are raised about an aspect of model performance (such as a revealed bias on a subclass of data \citep{buolamwini2018gender}), a new verifiable evaluation attestation can be created to examine the issue. This process is extremely costly, matching the cost levels of pre-emptive evaluation. An approach where a user challenges the performance of the model is possible, where a user provides a new test set to run on (as suggested by \citet{kang2022scaling}) requiring both a larger upfront cost and subsequent reward incentives. Such larger audits might be requested by civil liberty organizations or occur due to new standards (such as new bias requirements or evaluations in medical ML settings).

\section{Extended Background and Related Works} \label{sec:background}
The field of secure inference of ML models has grown rapidly from specialized interactive proof protocols~\citep{Ghodsi2017SafetyNetsVE} to more general purpose approaches, such as through multi-party computation~\citep{Knott2021CrypTenSM, Mishra2019DE} and ML inference using homomorphic encryption~\citep{Lou2021HEMETAH, juvekar2018gazelle}. These approaches, while preserving privacy of inputs, fail to provide publicly verifiable proof that ML inference occurred correctly.

Instead, much attention has turned to the inference of ML models inside zkSNARKs (Zero-Knowledge Succinct Non-Interactive Argument of Knowledge). Previous work in zero-knowledge machine learning (zkML)~\citep{Feng2021ZENAO} drew on older slower proving systems~\citep{Groth2016}, or using model architecture specific optimizations~\citep{Liu2021zkCNNZK, Lee2020vCNNVC} (such as tailoring for convolution neural networks). These limitations were addressed by \citet{kang2022scaling}, which followed ezkl by leveraging Plonkish arithmetization through the Halo2 \citep{halo2book} proving system. Recent work has also shown that inference proofs of specific LLMs are possible with custom cuda circuits \cite{sun2024zkllm}.

This work builds upon and heavily leverages the ezkl\footnote{\url{https://github.com/zkonduit/ezkl}} toolkit, which itself is underpinned by the Halo2 proving system. This toolkit is continually under development, targeting the deployment of verifiable execution of ML models on-chain in Web3 contexts. 
Although there is a related growing body of literature on verifiable training \citep{Sun2023zkDLEZ, garg2023experimenting, Jia2021ProofofLearningDA}, we do not use or address the issue of training here. Since the release of the first version of this manuscript, recent work has scaled zkSNARKs for small LLMs using the same underlying proving system (halo2) \citep{chen2024zkml} and ezkl itself \citep{ganescu2024trust}.

The idea of using zkML as a tool to create verifiable accuracy claims emerges naturally as an extension from previous zkML work. Previous works mention verifiable accuracy claims as applications of their zkML proof approaches, with \citet{kang2022scaling, chen2024zkml}, but do not explain the evaluation-focused threat models, aggregation options, or benchmarking system.
\citet{Weng2022pvCNNPA} takes this idea and extends it in a constrained multi-stakeholder environment context for CNNs using collaborative inference and homomorphic encryption, and recent work has shown similar trustless fairness benchmarking approaches using logistic regressions \citep{Tang2023PrivacyPreservingAT}.
This work differentiates from the above as the first to provide a generalizable framework for neural network benchmarking, going much further in identifying constraints to provide a practical framework for deployment, as well as discussing choices of evaluation data across context and model type. 

The idea of building accountability into algorithms like this dates back to \citet{kroll2015accountable}, highlighting the important role of cryptography in creating accountable approaches to computer science, a mandate we build upon here. Continuous and post-hoc auditing to ensure these systems live up to their claims \citep{kim2017auditing} and synchrony between the design of accountable systems and their regulation are essential for this to succeed \citep{desai2017trust}. 
In general this work builds on the broad and burgeoning fields of ML reproducibility \citep{Beam2020ChallengesTT,semmelrock2023reproducibility,Albertoni2023ReproducibilityOM,Belz2023MissingIU,10.5555/3546258.3546422,10.1001/jama.2019.20866,10.1145/3576915.3623130,gundersen2022machine} and AI fairness \citep{Muthukumar2018UnderstandingUG,Mehrabi2019ASO,CorbettDavies2018TheMA,May2019OnMS,Raji2019ActionableAI,Krkkinen2021FairFaceFA,Liu2023FairCompassOF,Parraga2023FairnessID}. 
While this literature seeks to ensure unbiased, reproducible, and fair ML models, it often does not address the practical risks that model providers will create unbiased and reproducible models and then later choose to deploy lower quality models due to cost, compute, or performance incentives.
We seek to pragmatically address this issue to give users the ability to verify that the model they are using actually meets relevant benchmarks of bias and fairness. 

\section{System Details}

\subsection{Flexible Model Setup}\label{sec:setup}

One of the key features is the flexible nature of this proof system. The ezkl toolkit, which underpins the system and is a contribution from the authors, takes an ONNX file as input, an open standard for saving ML models that can be accessed from standard ML libraries.

Each operation in the model's computational graph is represented by one of the ONNX operation types, and each operation will be converted into a proof constraint. While simple in concept, implementation details are critical to achieve efficiency. For example, most operations in the model can be reduced to Einstein summations to minimize the number of constraint implementations needed~\citep{danteHoneyBlog}.

These constraints are constructed into a large proving table for Halo2~\citep{halo2book}. Many further optimizations occur here to achieve speed improvements which are detailed elsewhere. Notably, the ability to create parallel region layouts using cycles, the use of fixed-column lookups to represent nonlinear functions, and the choice to perform operations using fixed-point arithmetic for speed~\citep{danteHoneyBlog}.

While generating this circuit for proving there are many calibration choices to be made around quantization and scale. These can be summarized as tradeoffs between accuracy and the quantity of resources needed for the proof. In this work, calibration for resources is preferred for large model sizes.

Beyond the proof circuit for inference of the model operations, we introduce an additional set of actions in the circuit to calculate a proof of the hash of the model weights (via ZKG hashing per \citet{ezkl2023zerooverheadhashing}). This results in a final proof that contains both the input-output values (which are made public in the proof circuit) and the model weight hash. 

The outputs of this step are the compiled circuit, a large proving key, and a small verification key. \textbf{This is the first public, open-source, working implementation of a zkSNARK proving system that can be used for any type of ML model.}

\subsection{Proof System and Arguments} \label{sec:proofsystem}

The flexibility of the ezkl proof system stems from the library's reduction of any zk-equivalent of an ML operation into some combination of four arguments: 
\begin{enumerate}
\item A cumulative dot product argument. 
\item A cumulative sum/product argument. 
\item An elementwise addition/multiplication/subtraction argument. 
\item A lookup argument, used to represent and constrain non-linear functions within the zero-knowledge-circuit. 
\end{enumerate}

These ``arguments'' are used to enforce and guarantee that the result of a computation is indeed from a particular agreed-upon computation graph (such as a neural network).

\subsubsection{Cumulative Arguments} \label{sec:cumarg}

Arguments 1. and 2. are constructed in a similar fashion. In the case of the cumulative sum/product consider a vector $\mathbf{x}$ of length $N$. To create a set of constraints within our zk-circuit we create a new vector $\mathbf{m}$ of length $N-1$ and set the following elementwise constraints: 
\begin{equation}
    x_i \circ m_{i} = m_{i + 1} \ \forall i \in 1..N, \ \text{where} \ m_0 = 0.
\end{equation}

$\circ$ is the addition operator in the case of the cumulative sum and the multiplication operator in the case of the cumulative product. The final element of the vector $\mathbf{m}$ then represents the cumulative sum or product, and the constraints above are enforced within the circuit using Halo2 \textit{selectors}. 

The dot product argument is constructed in a very similar fashion. We now have a second input vector $\mathbf{y}$, also of length $N$ and constrain the following: 
\begin{equation}
x_i \circ y_i + m_{i} = m_{i + 1} \ \forall i \in 1..N, \ \text{where} \ m_0 = 0.
\end{equation}

As before, the final element of the vector $\mathbf{m}$ then represents the dot product. 

\subsubsection{Elementwise Arguments} \label{sec:elemarg}

Argument 3. is constructed without leveraging the intermediate calculations of \ref{sec:cumarg}. Consider two vectors $\mathbf{x}$ and $\mathbf{y}$, both of length $N$. We constrain the resulting output $\mathbf{m}$ via the following: 
\begin{equation}
x_i \circ y_i = m_{i} \ \forall i \in 1..N, 
\end{equation}

\subsubsection{Lookup Arguments \label{sec:lookuparg}}
The ezkl system leverages halo2 as a proving backend with a few modifications. Most significantly, it changes the original lookup argument, typically used to constrain non-linear functions like ReLU, to the more efficient logUP lookup argument \citep{habock2022multivariate}.

\subsubsection{More Complex Arguments} \label{sec:complexarg}

More complex arguments such as those for the $\mathbf{min}$ and $\mathbf{max}$ functions, can be constructed as combinations of the above arguments. 
For instance the $\mathbf{max}$ argument can be constructed as follows: 

\begin{enumerate}
    \item Calculate the claimed $$m=\text{max}(x),$$ and instantiate a lookup table $\mathbf{a}$ which corresponds to the $\mathbf{ReLU}$ element-wise operation.
\item Constrain $\mathbf{w} = \mathbf{x} - (m - 1)$.
\item Use lookup $\mathbf{a}$ on  $\mathbf{w}$, this is equivalent to clipping negative values: $\mathbf{y}=\mathbf{ReLU}(w).$
\item Constrain the values $\mathbf{y}$ to be equal to 0 or 1, i.e. use a selector that enforces that $$y_i*(y_i - 1)=0, \qquad \forall i \in 1\dots N.$$
Any non-clipped values should be equal to 1 as at most we are subtracting the max.
This demonstrates that the there is no value of $\mathbf{x}$ that is larger than the claimed maximum of $\mathbf{x}$.
\item We have now constrained $m$ to be larger than any value of  $\mathbf{x}$, we must now demonstrate that \textit{at least one} value of  $\mathbf{x}$ is equal to $m$, i.e that $m$ is an element of  $\mathbf{x}$. We do this by constructing the argument $$\mathbf{x} = \mathbf{ReLU}(1 - \sum_i y_i) = 0.$$  
Note that $$\sum_i y_i = 0 \iff z = 1$$ and thus no values of the witness are equal to $$\text{max}(x).$$ 
Conversely, if $$\sum_i y_i >= 1 \iff z=0$$ and thus least one value is equal to 1.  
\end{enumerate}

\subsection{Dataset Inference} \label{sec:datasetinference}

The most computationally expensive and decision-intensive step is the inference of the ML model over the benchmarking dataset to generate the zkSNARKs. One extremely key choice is the selection of a relevant dataset, an issue we leave for \autoref{sec:datasetchoice}. Instead, we focus here on the mechanics of how the inferences can be done.

While zero-knowledge ML is often described in terms of the inference occurring inside the proof generation, it can more aptly be described as the proof verifying that an inference was performed; a distinction that has important practical implications.

Firstly, the model can be run over the dataset inputs as it would in any other inference context to generate input-output pairs, $(x_i,y_i)$. These pairs are then quantized per the setup choice above to produce the witness inputs,  $(\tilde{x}_i, \tilde{y}_i)$. When accuracy is sacrificed, the quantized input-output pair may be different from the original values by a few percent (this can be calibrated). Importantly, as far as the test set is concerned, this accuracy trade-off can be examined \emph{before} performing the expensive proof step. The proof step will verify that the witness could be authentically generated using the model weights (which are treated as private inputs for the proof). Hence, iterative calibration on the benchmarking test set to ensure witness data has the performance characteristics is important.

For each of these witness files, a proof is generated using the proving key and the circuit. This is the slowest part of the stack and scales linearly with the size of the test set. For each witness file, $(\tilde{x}_i, \tilde{y}_i, H(W))$, a proof file will be generated $\Pi_i$. These proof files are each very small (on the order of kilobytes) and can be individually verified using the verification key or aggregated.

\subsubsection{Posting Performance Claims}
Other work has discussed the use of specific billboards for sharing attestations \citep{Tang2023PrivacyPreservingAT}, but the portable nature of the verifiable evaluation attestations is that they can be shared or hosted anywhere and copied repeatedly without additional risk. As such, posting performance claims to Github, Papers with Code, Huggingface, IPFS, another third-party public-facing hosting service, or even one's own website is sufficient. Ideally, this posting would then be mirrored elsewhere for accountability. 

\subsubsection{Hardware}
All experiments run here can be done on commodity hardware. They were initially all done on a 10-core Intel Xeon Processor E5-2687W with 1T of RAM. Some parts of the proving stack (including those using Halo2) are parallelized across CPU cores. The key hardware constraint is the RAM size for storing the proving key, hence the choice of this machine. GPU acceleration is possible in parts of the proving system (and has been done end-to-end for LLMs with custom Cuda circuits recently \citep{sun2024zkllm}), but no experiments here used a GPU. Experiments in the final version of this paper were run on a cloud compute cluster provided by ezkl to customers, with similar hardware to the above, but was able to achieve 20\% faster proof times than the Xeon above.

\subsection{Scalability \& Future Speed Improvements} \label{sec:speedup}

One of the main constraints on the application of this system (especially with regard to large foundation models) is the speed with which proofs can be generated. This work builds on the ezkl toolkit, which is constantly undergoing speed improvements through optimizations to circuits for inference operations and engineering improvements. Future work will improve these speeds through proof splitting and parallelization \citep{ezkl2023splitting}, GPU acceleration \citep{ezkl2023gpu, derei2023accelerating}, or using alternative underlying proof systems \citep{kothapalli2022nova, boneh2020halo, setty2020spartan}.
Other approaches such as cqlin from \citet{cryptoeprint:2023/393} show promise for ML, while other unpublished work has performed further optimizations \citep{kang_2023_tensorplonk}. GPUs acceleration through implementing attention circuits in cuda has proven effective in creating significant speed improvements for LLMs \citep{sun2024zkllm} and recent work has shown the inference of small LLMs in zkSNARKS \citep{ganescu2024trust, chen2024zkml}.

Interestingly, proof splitting may prove extremely exciting as future work. As we see in \autoref{sec:increasingsize}, the time and resource complexity of models is sublinear. As a result, splitting an AI model into chunks (e.g., each attention block) and completing a proof for each chuck should provide a lower overall computational cost. It's possible that even the largest models could be chunked into reasonable sized parts that could be proved with current hardware resources. 

Further, it is possible to optimize the design of models for more efficient inference in zkSNARKs \citep{Jha2023DeepReShapeRN}. Similarly, choices can be made during benchmarking design, such as model inference batching, which can have small speed improvements at the cost of larger proof requirements. 

As we see improvements in the speed of underlying proving systems and their hardware, the sublinear growth of proving time means that foundation models (which are increasingly performant at small sizes) will be commercially viable at scale.


\end{document}